\documentclass[10pt,twocolumn,letterpaper]{article}

\usepackage{iccv}
\usepackage{times}
\usepackage{epsfig}
\usepackage{graphicx}
\usepackage{amsmath}
\usepackage{amssymb}
\usepackage{subfig}
\usepackage{booktabs}
\usepackage{comment}
\usepackage{adjustbox}

\usepackage[breaklinks=true,bookmarks=false]{hyperref}

\iccvfinalcopy 

\def\iccvPaperID{****} 

\ificcvfinal\fi

\begin{document}

\title{DECOMPL: Decompositional Learning with Attention Pooling for Group Activity Recognition from a Single Volleyball Image}

\author{Berker Demirel\\
Sabanci University\\
{\tt\small berkerdemirel@sabanciuniv.edu}
\and
Huseyin Ozkan\\
Sabanci University\\
{\tt\small hozkan@sabanciuniv.edu}
}

\maketitle
\ificcvfinal\thispagestyle{empty}\fi

\begin{abstract}
   Group Activity Recognition (GAR) aims to detect the activity performed by multiple actors in a scene. Prior works model the spatio-temporal features based on the RGB, optical flow or keypoint data types. However, using both the temporality and these data types altogether increase the computational complexity significantly. Our hypothesis is that by only using the RGB data without temporality, the performance can be maintained with a negligible loss in accuracy. To that end, we propose a novel GAR technique for volleyball videos, DECOMPL, which consists of two complementary branches. In the visual branch, it extracts the features using attention pooling in a selective way. In the coordinate branch, it considers the current configuration of the actors and extracts the spatial information from the box coordinates. Moreover, we analyzed the Volleyball dataset that the recent literature is mostly based on, and realized that its labeling scheme degrades the group concept in the activities to the level of individual actors. We manually reannotated the dataset in a systematic manner for emphasizing the group concept. Experimental results on the Volleyball as well as Collective Activity (from another domain, i.e., not volleyball) datasets demonstrated the effectiveness of the proposed model DECOMPL, which delivered the best/second best GAR performance with the reannotations/original annotations among the comparable state-of-the-art techniques. Our code, results and new annotations will be made available through GitHub after the revision process.
\end{abstract}

\section{Introduction}

Group activity recognition (GAR) refers to the identification of the collective activity performed by a group of individuals in a given short video clip \cite{choi2009they, bagautdinov2017social, gavrilyuk2020actor, ibrahim2016hierarchical, qi2018stagnet, wu2019learning, yan2018participation, yuan2021learning}. For the goal of GAR, unlike action recognition \cite{moeslund2006survey, ahmad2006human, fihl2006action}, one should jointly consider multiple individual actions that are statistically dependent both spatially and temporally. Even when the GAR problem can be reduced to finding the most critical actor's (i.e., individual's) action, that action typically depends on the configuration as well as the actions of other actors in the  scene. Hence, considering the individual actions jointly and extracting the higher level semantic information is important. Such higher level representations appear as a key component in several other areas as well, e.g., social behavior analysis \cite{nabi2013temporal}, surveillance systems \cite{lao2009automatic}, sports video analysis \cite{direkoglu2012team, qi2018stagnet}, social robots \cite{nan2019human} and even autonomous driving \cite{braunagel2015driver}. The GAR studies can benefit from the ideas and approaches developed in these areas. However, exclusively in the spatio-temporal GAR problems, it is challenging to efficiently build up the spatial as well as temporal relations among individuals based solely on the processing of videos that are short in time.

\begin{figure}[!t]
    \centering
    \includegraphics[  width=9cm,
  height=6.5cm,
  keepaspectratio]{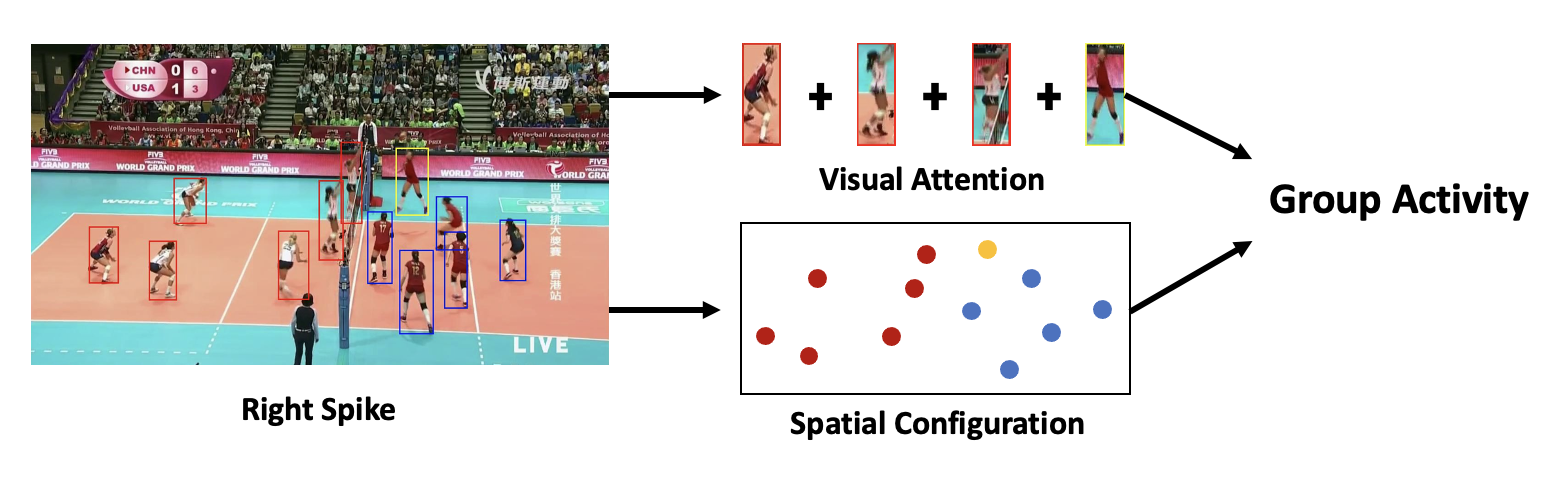}
    \vspace*{-6mm}
    \caption{Group activity recognition using RGB image and bounding box coordinates. Visual representation is captured by the attention mechanism and the complementary position features are modelled with the proposed coordinate block.}
    \label{fig:mini_figure}
\end{figure}
In this paper, we propose a novel technique, called DECOMPL, which tackles the GAR problem for volleyball videos only in the spatial dimension. Changing the problem setting from spatio-temporal to only spatial has its own advantages and limitations. It eases the computation by a significant margin and reduces the problem into discovering the relations of individuals in the input image. On the other hand, solving the problem with less information is obviously more challenging. Although the pose (of an individual) and the spatial configuration of poses are static in a given image and has no temporal dimension, it is known to be highly predictive of the corresponding action in time \cite{moeslund2006survey, fihl2006action}. Here we hypothesize that, conditioned on the spatial information, dropping the temporal dimension in favor of computation gains should not be losing much information regarding GAR in volleyball videos.

Our technique (Figure \ref{fig:mini_figure}), DECOMPL, consists of two information processing branches, the visual branch and the coordinate branch, for extracting person\footnote{We use the words ``individual", ``actor", ``player" or ``person" interchangeably with slight contextual differences.} level visual features (encoding the individual actions based on their static poses) and person level spatial location features (encoding the spatial configuration of the individuals). 
On the visual branch, we incorporate the VGG backbone \cite{simonyan2014very} with RoI align \cite{he2017mask} to extract the person level features from an image. A multi-head attention pooling module \cite{ilse2018attention} also assigns importance weights to the extracted person level features. As for the other branch, we use i) a coordinate module to extract the coordinates and the corresponding spatial location features for each individual, and ii) a single attention pooling module to combine the extracted coordinate based features with respect to their attention weights. The information flowing over these two branches is aggregated for each image, and a second aggregation across images reveals our final GAR decision for the short video clip in hand. In this GAR process, DECOMPL exploits the decomposable structure in the volleyball videos, thanks to the mirror symmetry across the team sides, by introducing certain sub-classification tasks. In addition to deciding on the group activity label, we also decide which team side does the corresponding activity and what kind of activity has been seen without the side information. With these sub-classification tasks, DECOMPL achieves to reinforce the supervision loss signal and increases its representation capacity.

\begin{figure*}[!ht]
    \centering
    \includegraphics[  width=20cm,
  height=6.5cm,
  keepaspectratio]{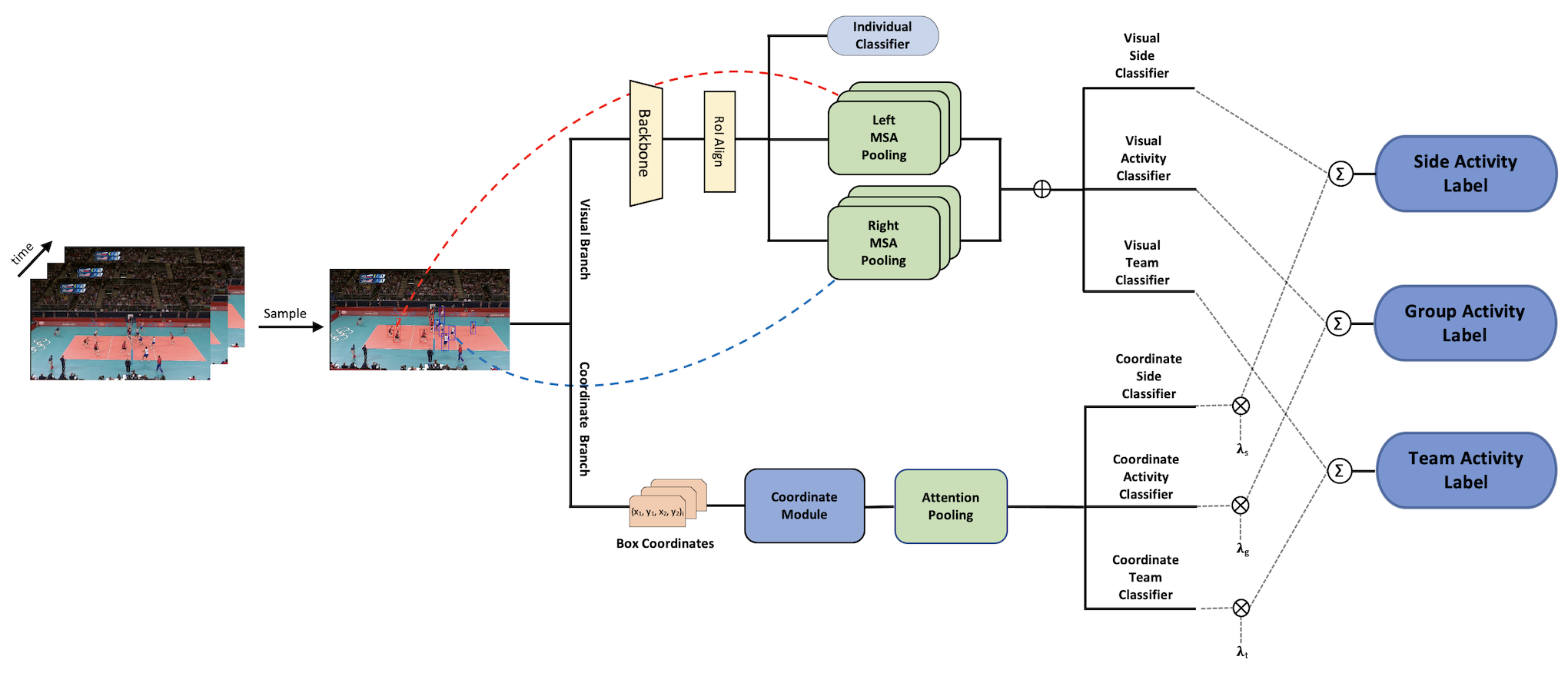}
    \vspace*{-4mm}
    \caption{Overview of DECOMPL that consists of two main branches. On the visual branch, RGB features are extracted using VGG-backbone and RoI align. Features of the two teams are separated and passed to multi-head attention pooling mechanisms and concatenated to represent the visual features of the scene. On the coordinate branch, box coordinates are passed to the coordinate block that models the interactions between players and they are summarized by an attention pooling. Scene representations from each branch decides for the side activity, group activity and team activity and their inferences are fused by learnable parameters $\lambda_s$, $\lambda_g$ and $\lambda_t$. Best viewed in color.}
    \label{fig:model_figure}
\end{figure*}

Our main contributions and highlights are as follows.
\begin{itemize}
    \item DECOMPL utilizes MIL pooling \cite{ilse2018attention} with multi heads for the GAR problem as the first time in the literature. Unlike other state-of-the-art methods \cite{yuan2021spatio, zhou2021composer}, DECOMPL is end-to-end.
    \item We decompose the GAR for volleyball into subproblems by exploiting the mirror symmetry across the team sides and introduce auxiliary labels. Benefiting from the extra information, representation capacity of the model is improved.
    \item By dropping the temporality, we reduced the number of required floating point operations to $10\%$ of the nearest competitor, with only negligible loss (less than $0.3\%$) on the accuracy side. 
    \item In our extensive experiments and ablation studies with the widely used benchmark Volleyball dataset (VD) in \cite{ibrahim2016hierarchical}, DECOMPL achieves $93.8\%$ GAR performance (the second highest) for the original VD, and $95.2\%$ (the highest) for the reannotated version. Our coordinate branch (using only the configuration of players) single-handedly performs up to $73\%$ accuracy without any visual information, which is on par with most of the early deep learning based solutions.
    \item Although this VD is very popular in the literature, we encountered systematic labeling flaws in its ground truth. We manually reannotated ($497$ examples were corrected out of $4821$), discussed the semantics of each label and reported the results for the methods in \cite{yuan2021spatio,zhou2021composer, li2021groupformer} with the new annotations. 
\end{itemize}

\section{Related Work}

Early GAR methods typically used convolutional neural networks (CNN) and input the CNN features to recurrent neural networks (RNN), several examples include \cite{ibrahim2016hierarchical, shu2017cern, bagautdinov2017social, qi2018stagnet, ibrahim2018hierarchical, hu2020progressive}. Newer approaches like relational modules and graph CNNs have been used to capture the group level information more powerfully \cite{ibrahim2018hierarchical, azar2019convolutional, wu2019learning, hu2020progressive, pramono2020empowering}. Recently, attention models, particularly transformers, have been utilized to detect the most important actors in the scene \cite{qi2018stagnet, yuan2021learning, gavrilyuk2020actor, li2021groupformer}. We observe that these GAR methods can be grouped into three categories with respect to the input they use: RGB only, keypoint only and mixed. 

RGB only methods take only an RGB image as input. A 2-stage deep hierarchical model that utilizes LSTMs was used in the RGB method of \cite{ibrahim2016hierarchical} to form representations at the temporal level for individuals as well as groups. Likewise, \cite{shu2017cern} also used 2-level LSTMs, but minimizes an energy score to get the group activity predictions. In \cite{bagautdinov2017social}, the predictions were based on a fully-convolutional network module that generates multiscale features (with resizing procedures) that are fed to an RNN to model temporality. Certain other methods have utilized graph convolutional networks (GCNs) to infer relations among individuals \cite{ibrahim2018hierarchical, azar2019convolutional, wu2019learning, hu2020progressive, pramono2020empowering}. GCNs in \cite{wu2019learning} created an actor-relation graph to simultaneously combine relations between spatial and visual features. Similarly, \cite{hu2020progressive} created a graphical representation and combines with state, action and reward ideas inspired by reinforcement learning. Further, \cite{ehsanpour2020joint} found the most critical actor by using a self-attention module. It used an additional graph attention to model the relational information among agents with an I3D backbone \cite{carreira2017quo}, and capture the temporal context. Spatial features were discovered in \cite{yuan2021spatio} with a CNN backbone and processed via dynamic relation and dynamic walk reasoning modules.

Keypoint only methods use coordinate-based keypoint representation of the actors, and also the ball trajectories \cite{zappardino2021learning, gavrilyuk2020actor, thilakarathne2021pose, perez2022skeleton, zhou2021composer}. The method in \cite{zappardino2021learning} directly used the spatial coordinates of actor joints, and created a relation module and an attention mechanism to describe the image frame with a single feature vector. In addition to the pose skeleton, the ball tracklets were utilized in \cite{perez2022skeleton} to learn the interactions between individuals. \cite{zhou2021composer} proposed to use a multiscale transformer to perform compositional learning from tokens of ball tracklets and keypoint coordinates. 

Finally, mixed methods typically use multiples of RGB, optical flow, and keypoint inputs \cite{azar2019convolutional, pramono2020empowering, yuan2021learning, li2021groupformer} together. Spatial and temporal features were extracted in \cite{pramono2020empowering} by using the self-attention mechanism and then utilized as a conditional random field. A clustered spatio-temporal transformer can also be effectively used as an encoder-decoder mechanism for building the relations across features \cite{li2021groupformer}.

Our method falls in the category of RGB only methods. We utilize a single RGB image per video clip during training; whereas in the testing phase, we obtain the classification output through a simple averaging of the independent decisions from frames of the test video clip. Since the final decision does not give weight to any particular frame, and since there is no joint consideration of frames, our method does not exploit temporal modeling. Moreover, we do not use any optical flow, keypoint, or pose information. For these reasons, our method is computationally much simpler and faster in run time compared to the other methods, while providing a GAR accuracy that is on par with the highest performance figures reported in the literature.

\section{Method}

The proposed technique DECOMPL is demonstrated in Fig. \ref{fig:model_figure}. The task is to recognize the group activity in a short volleyball video clip that is given together with the box coordinates of the players. During testing, our algorithm produces GAR decisions for each frame first, and then an aggregation across all frames yield the final GAR decision. In the training phase, GAR decisions are made based on individual frames that are uniformly sampled from short video clips, with each decision based solely on the visual and coordinate features of the selected frame. DECOMPL splits the task into two different branches, the visual and the coordinate branches. 

In the visual branch, a VGG backbone is used to extract the visual features of the frame, from which the individual level features are obtained via the RoI align \cite{he2017mask} and projected onto a $D$-dimensional space of $X \in R^{N \times D}$. Here, $N=12$ is the number of players. To capture the team level features, we sort the feature list with respect to the x-coordinates of the boxes players are in and obtain two sublists: left team features and right team features. These team level features are fed into the multi-head attention pooling network (Section \ref{sub:AP}) to get $X_{l} \in R^{D}$ and $X_{r} \in R^{D}$. The frame level features $X_{visual}$ (output of the visual branch) is obtained by concatenating $X_{l}$ and $X_{r}$.

In the coordinate branch, only the bounding boxes are processed to furnish the model with the configuration of the players in the scene. Since the player configuration is independent of the visual features, these two branches can be computed concurrently. The box coordinates of the players, $X_{b} \in R^{N \times 4}$, are fed into our coordinate module to capture the distance relations of the players. We use the same embedding dimension to project the location features onto $X_{loc} \in R^{N \times D}$ and the embedded location features are then pooled using the attention pooling (Section \ref{sub:AP}). As we only want to represent the relative positions of the players in the input frame, features are not split before the pooling. Hence, the coordinate feature vector $X_{coordinate} \in R^{D}$ is passed as output directly.

\subsection{Attention Pooling}\label{sub:AP}
Since a scene summary should reflect the significant players' features to a larger degree, we consider that a weighted pooling of the player features is necessary for a compact representation. The max pooling is not appropriate as it eliminates the higher level semantic information and applies hard selection in an index-wise fashion. The mean pooling cannot weigh the important actors suitably, ending up with a summary that has a low signal to noise ratio. Therefore, simple poolings like max and mean operators are too naive to obtain the necessary information from the frame. Further, since it is hard to order the players on the court plane (a parsing issue), it is important to achieve permutation invariance in the weighting scheme. In order to solve these issues, we adopt the attention pooling mechanism of \cite{ilse2018attention} which is trainable, permutation invariant, and which can assign weights to the players with respect to their contributions (importances) to the final GAR accuracy.

Given the input of embeddings $X = \{x_1, ... , x_N\}$, the process of attention pooling can be formulated as:

\begin{equation}
X_{\text{pooled}} = \sum_{i=1}^{N} a_i  x_i,
\end{equation}
\noindent
where

\begin{equation}
a_{i} = \frac{\exp \{  w^{\top} \tanh(V x_{i}^{\top}) \}}{\sum_{j=1}^{N} \exp \{  w^{\top} \tanh(V x_{j}^{\top}) \}}.
\end{equation}
\noindent
Here, $w \in R^{L \times 1}$ and $V \in R^{L \times D}$ where $L$ is the hidden dimension and $\tanh(.)$ is the non-linear hyperbolic tangent. Our algorithm uses the attention pooling as is on the coordinate branch. For the visual branch, we extend this approach to multiple heads; and the output of each pooling head is stacked and projected onto the original dimension.

\subsection{Coordinate Module}
Our coordinate module takes the box coordinates $X_{b} \in R^{N \times 4}$ as input and first sorts them from left to right. Then, the pairwise difference vectors are generated for each individual $X_{pd} \in R^{N \times N \times 4}$. Each difference vector is projected onto a real number by sharing the weights to extract their value in the current configuration. For that, we apply a convolution kernel of size $4$ with stride $4$. Considering closer players have more in common in terms of the information that they supply to the configuration, the coordinate module convolves the features to get semantically higher level information. Output of the convolutional layers are in the end projected onto the D-dimensional space to get coordinate features $X_{c} \in R^{N \times D}$.   

\subsection{Multiple Loss Signals}
There are mainly two losses that guide in the volleyball activity recognition: activity loss and individual loss \cite{azar2019convolutional, bagautdinov2017social, ibrahim2018hierarchical, ibrahim2016hierarchical, qi2018stagnet, shu2017cern}. In DECOMPL, we utilize the symmetric property of the label structure and introduce auxiliary labels by decomposing the original ones. Each clip has its own side label (left / right), sideless team activity (pass, win, set, spike) and side-sensitive group activity labels (left spike, right win) as well as individual activity labels (setting, blocking). Therefore, it is possible to make the output of the visual and coordinate branches more representative. Individual activity labels are inferred by only using the embedded vector after RoI align whereas the other labels are obtained through the inputs $X_{coordinate}$ and $X_{visual}$. Then, side, sideless team activity and side-sensitive group activity decisions of the visual classifiers are fused with the ones of coordinate classifiers, using the learnable parameters $\lambda_s$, $\lambda_g$, and $\lambda_t$. This allows our model to incorporate the configuration information with the visuals. Note that without the visual information, the model cannot distinguish players; whereas without the coordinate information, the model is agnostic to the relative positions. Hence, both are required for an effective solution.

Our end-to-end network is optimized with the total loss
\begin{equation}
\mathcal{L}_{\text{total}} = \mathcal{L}_{\text{individual}} + \mathcal{L}_{\text{group}} + \beta (\mathcal{L}_{\text{side}} + \mathcal{L}_{\text{team}}),
\end{equation}
\noindent 
where $\beta$ is a hyperparameter.

\begin{figure*} [!ht]
    \centering
  \subfloat[\label{fig1:1a}]{%
       \includegraphics[width=7cm,
  height=6.5cm,
  keepaspectratio]{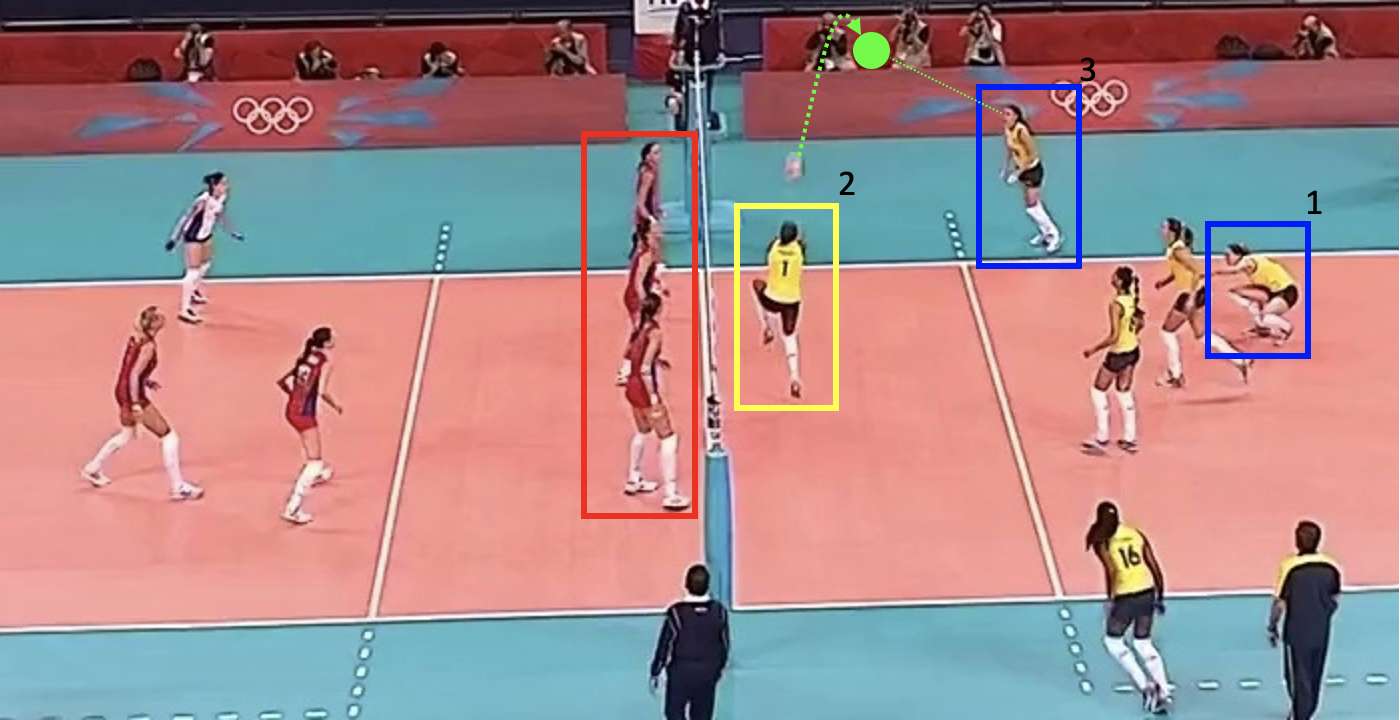}}
    \hfill
  \subfloat[\label{fig1:1b}]{%
        \includegraphics[width=7cm,
  height=6.5cm,
  keepaspectratio]{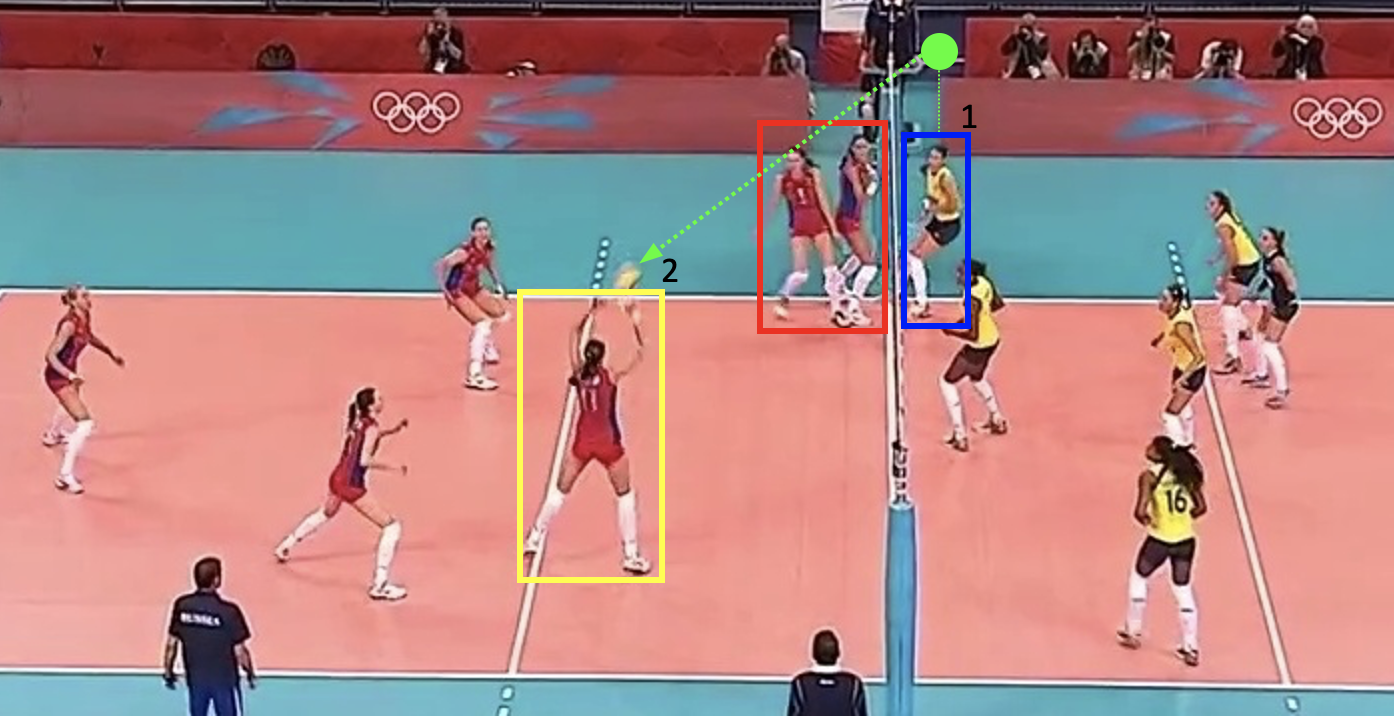}}
  
  \caption{Some examples of flawed labeling from the original Volleyball dataset \cite{ibrahim2016hierarchical}. Indices represent the order of action performed by a player. The main players are indicated with yellow bounding boxes. (a) is annotated as right-pass activity while the true annotation is right-set. (b) is an instance of left-set activity while the true annotation is left-pass.}
  \label{fig1} 
\end{figure*}

\section{Volleyball Dataset (VD): Reannotations}
In our performance evaluations, we conducted extensive experiments with the VD of \cite{ibrahim2016hierarchical} which is also widely used for GAR in the literature and publicly available. This dataset originally contains 55 volleyball videos with $4830$ labeled frames ($3493$ / $1337$ for training / testing). Each clip is labeled with one of the $8$ side-sensitive group activity categories: right set, right spike, right pass, right win-point, left set, left spike, left pass and left win-point. Moreover, the centered frame in each clip is annotated with $9$ individual action labels: waiting, setting, digging, falling, spiking, blocking, jumping, moving and standing. However, when looked into carefully, one can see that the train and test sets contain some outliers. The dataset has a few video clips whose point of view is not a regular horizontal perspective. Despite that this dataset is very popular and has been used in a number of previous studies, e.g., \cite{yuan2021spatio, azar2019convolutional, li2021groupformer}, we realize that there exist falsely labeled clips. Also, adopting a different labeling approach may result in a more useful dataset for group activity recognition.

Particularly, we observe that clips were labeled -especially the ``set'' and ``pass'' examples- based mostly on the pose of the main acting player in the scene. For instance, a clip was labeled as ``pass'' if the main player is bumping no matter where the ball goes after her/him. However, while annotating, the position of the ball after an action must be considered as well to emphasize the group activity. Similarly, a player being in an overhand pass pose might not necessarily mean that the activity is set. These activities have higher level semantic meanings that must be taken into account while labeling rather than looking solely into the pose of a player. The labeling scheme in VD may reduce the problem to pose estimation which might degrade the ``group'' concept in the activity detection. Therefore, we decided to reannotate the data, as another contribution of the presented study. 

According to our interpretation, the group activity label ``spike'' means that the ball passes to the other side in a ``comfortable'' position. It should be comfortable in order to distinguish a ``spike'' from a ``pass'' because, in quite a few examples in VD, the ball is forwarded to the other side just to save the point. This is mostly observed in situations where the defending team has difficulties in defending. If the ball is passed to the other side in a defensive manner, we reannotated it as ``pass''. We also reannotated a clip as ``pass'' if the player who is acting touches the ball with only defensive intention or s/he is the first player who touches the ball after the last touch of the opponent team. In this way, we take both the semantic meaning and the group concept back to the annotations. To annotate a clip as ``set'', we followed if the main actor in the scene is forwarding the ball to the one who is going to spike it. Lastly, ``win-point'' and left/right discrimination have no flaw in the original annotations; but there are random labeling errors as well which we corrected. The number of changes in our reannotations and the new statistics can be observed from Table \ref{data_stats}.

\begin{table}[ht]
\centering
\begin{tabular}{|l|l|l|}
\hline
Group Activity Class & Before & After    \\ \hline
Right set            & 644 & 596             \\ \hline
Right spike          & 623 & 640             \\ \hline
Right pass           & 801 & 830             \\ \hline
Right win-point      & 295 & 297             \\ \hline
Left set             & 633 & 605             \\ \hline
Left spike           & 642 & 654             \\ \hline
Left pass            & 826 & 831             \\ \hline
Left win-point       & 367 & 368             \\ \hline
\end{tabular}
\vspace{1.5mm}
\caption{Distribution of the group activity labels before and after reannotations.}
\label{data_stats}
\end{table}

\begin{figure*} [!ht]
    \centering
  \subfloat[\label{fig2:2a}]{%
       \includegraphics[width=7cm,
  height=6.5cm,
  keepaspectratio]{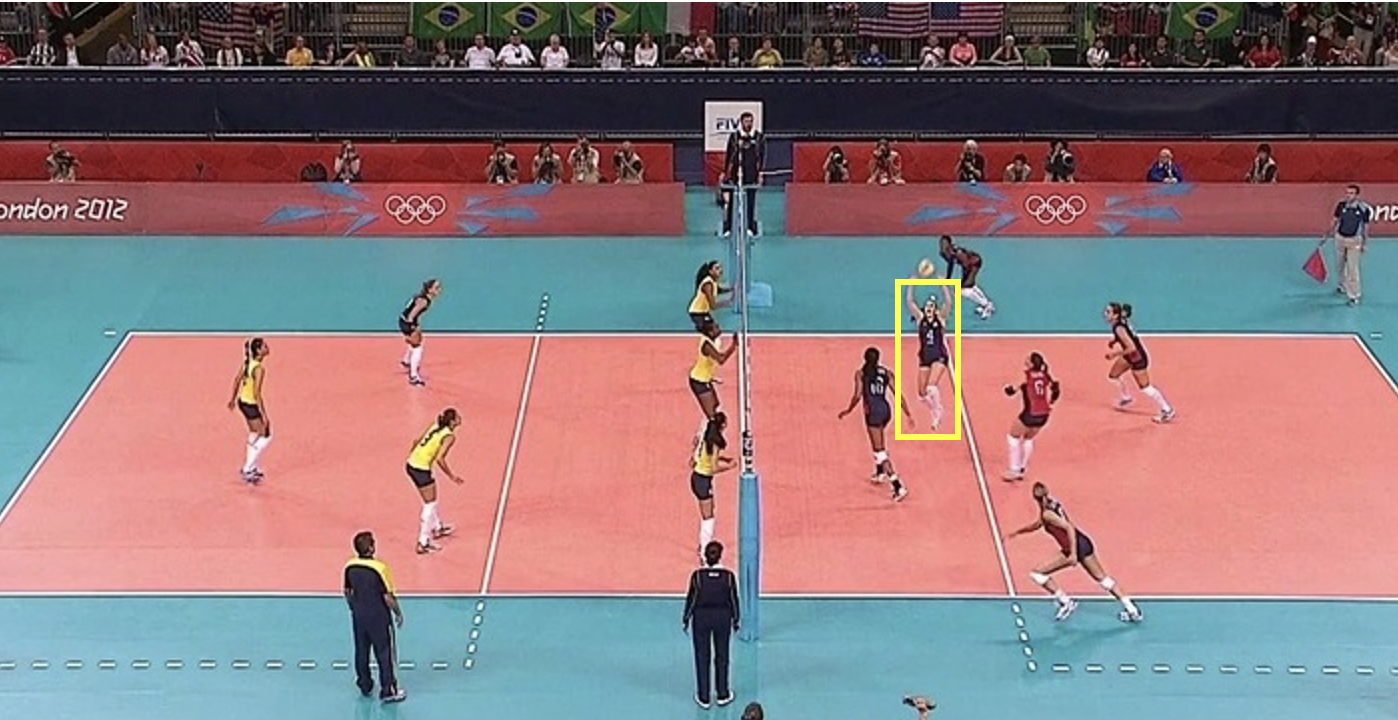}}
    \hfill
  \subfloat[\label{fig2:2b}]{%
        \includegraphics[width=7cm,
  height=6.5cm,
  keepaspectratio]{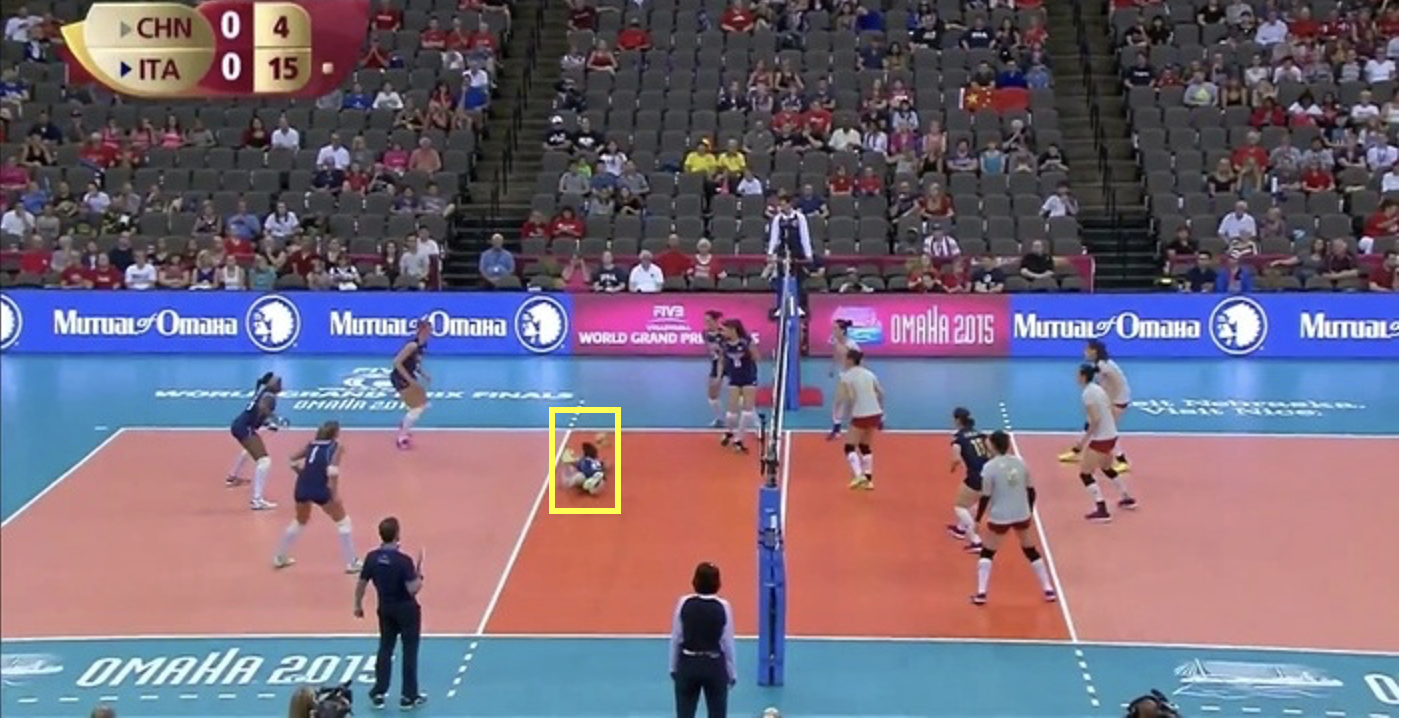}}
  \caption{Some examples of random errors from the original Volleyball dataset \cite{ibrahim2016hierarchical}. The main actors are indicated with yellow bounding boxes. (a) is annotated as left-set activity while the true annotation is right-set. (b) is an instance of right-pass activity while the true annotation is left-pass.}
  \label{fig2} 
\end{figure*}

After removing $9$ video clips due to the change in the camera angle, the refined dataset has $4821$ clips. We performed a total of $497$ reannotations which is approximately $\%10$ of the whole dataset. $381$ out of $497$ reannotations were a result of the deviation between the way we reannotated and the original annotations. We believe that our reannotations are more meaningful and useful from the GAR perspective, as described above. The residual $116$ reannotations, which is approximately $\%2.4$, were due to the random labeling errors in the original VD. For example, $29$ of these random errors are coming from side mismatch (i.e. when the activity is on the left side but the label says right or vice versa). Errors like these impede the training process of a statistical learner, and also its performance evaluation.

We provide visual examples in the following to illustrate the scheme that we applied in our reannotations and the random labeling errors in the original VD that we corrected.

\subsection{Examples of Our Reannotation Scheme}
Our reannotation scheme systematically deviates from approximately the $\%10$ fraction of the original annotations. We believe that for those examples, diverging from the original labeling process increases the quality of the dataset by representing instances more accurately. These are typically the labels that claim ``set" rather than ``pass" or vice versa. The following examples are gathered from three different matches and annotation differences like these can be found in all matches.

Fig. \ref{fig1:1a} can be seen as an example of an original annotation ``pass" where our annotation is rather a ``set". It can be seen from the figure that the common theme is setting a position to the spikers. The way the opponent team prepares to defend with blocks and the pose of the potential spikers are indicators of that theme. Therefore, we cannot decide just by the bumping pose of the actor but need to evaluate the scene as a ``group''.

On the other hand, Fig. \ref{fig1:1b} was originally labeled as ``set" instead of our labeling choice ``pass". If one investigates the image, s/he  observes that there is a player who has hit the ball and now descending and the actor's team has blockers in front of that player. This gives us the information that the opponent team has spiked and the actor is meeting the ball. Since the actor is the first player who touches the ball after the opponent team, we should label these images as ``pass".

\subsection{Examples of the Corrected Random Errors}
Random errors in the original VD generally seem to be due to the lack of attention of the annotator. We corrected these errors which include annotating frames as ``right" instead of ``left", ``spike" instead of ``pass" etc. None of such error types are dominant to the others. Namely, one might not expect that the number of times the annotation is mistakenly given as ``spike" instead of ``pass" significantly exceeds the number of times that is given as ``set" instead of ``spike". Thus, these errors are not systematic.

In Fig. \ref{fig2:2a} and Fig. \ref{fig2:2b}, it is clear that the side information is mistakenly annotated.

\section{Experiments}

In this section, we provide a detailed description of the Collective Activity dataset (CAD), present the implementation details of DECOMPL, and evaluate its performance by comparing it with state-of-the-art methods from the literature. Our results are based on the Volleyball dataset \cite{ibrahim2016hierarchical} and Collective Activity dataset \cite{choi2009they}, and the results for VD include both the original and corrected annotations. Moreover, an ablation study is presented to demonstrate the performance contribution of each block in our model.

\subsection{The Collective Activity Dataset}
The dataset contains a total of 2511 clips extracted from 44 short video sequences, featuring 5 different collective activities: crossing, walking, waiting, talking, and queueing. For each clip, the centered frame is labeled with bounding boxes and respective individual action classes which can be N/A or one of the five different activities. To align with prior research, the activity labels for ``walking'' and ``crossing'' are merged into a single category called ``moving''.

\subsection{Implementation Details}
We use PyTorch \cite{paszke2019pytorch} for the implementation, and follow the prior works \cite{bagautdinov2017social, yuan2021spatio} for extracting the annotations and reading the images. We resize images to $720\times1280$ for the VD and $360\times640$ for the CAD, and use horizontal flip augmentation. We also use several ($T=10$) successive images from the same video following the prior works. However, this is only for additional augmentation since our method does not rely on temporality as it uniformly samples a single frame from a given time window. VGG-19 \cite{simonyan2014very} backbone is used with RoI align \cite{he2017mask} (crop size = $4\times4$) to extract visual features of the actors with a dimension of $D=128$. Due to the additional labels and team structure used in the VD, the model architecture differs slightly between the VD and CAD. For the VD, we use the attention pooling of $2$ heads for the visual branch and $1$ head for the coordinate branch, while for the CAD we used a single head attention pooling mechanism for the visual branch. We use hidden dimension of $512$ for all attention modules. Due to the difference in the number of classification tasks between the datasets, our model for the VD has $3$ classification heads while the model for the CAD has only $1$. All classification heads are linear projections onto the label dimension in the corresponding dataset. The losses from different classification tasks are combined with $\beta = 1$ for the VD. For both datasets, we use ADAM optimizer \cite{kingma2014adam} with a learning rate of $0.0001$ which drops by a factor of $2$ every $30$ epochs, for a total of $120$ epochs. All experiments are conducted on $2$ RTX $3090$ GPUs with a batch size of $8$.

\begin{table}[ht]
\centering
\begin{adjustbox}{width=1\columnwidth}
\begin{tabular}{*5l} 
    \toprule
    Model & Input & Backbone & VD & CAD \\ 
    \midrule
    HDTM \cite{ibrahim2016hierarchical}  & RGB & AlexNet & 81.9 & 81.5 \\
    CERN  \cite{shu2017cern} & RGB  & VGG-16 & 83.3 & 87.2 \\
    stagNet \cite{qi2018stagnet} & RGB  & VGG-16 & 89.3 & 89.1 \\
    RCRG \cite{ibrahim2018hierarchical} & RGB  & VGG-16  & 89.5 & -\\
    SSU \cite{bagautdinov2017social} & RGB   & Inception-v3 & 90.6 & -\\
    SACRF \cite{pramono2020empowering} & RGB & ResNet-18 & 90.7 & 94.6  \\
    PRL \cite{hu2020progressive} & RGB  & VGG-16 & 91.4 & -  \\
    Ehsanpour \cite{ehsanpour2020joint} & RGB & I3D & 93.1 & 89.4\\
    ARG \cite{wu2019learning} & RGB  & Inception-v3 & 92.5 & 91.0\\
    HiGCIN \cite{yan2020higcin} & RGB & ResNet-18 & 91.5 & 93.4\\
    DIN \cite{yuan2021spatio} & RGB & VGG-16 & 93.6 & -\\
    GroupFormer \cite{li2021groupformer} & RGB & Inception-v3 & \textbf{94.1} & 93.6 \\
    
    \midrule
    Zappardino \cite{zappardino2021learning} & Keypoint & OpenPose  & 91.0 & -\\
    GIRN \cite{perez2022skeleton} & Keypoint & OpenPose  & 92.2 & -\\
    AT \cite{gavrilyuk2020actor} & Keypoint & HRNet  & 92.3 & - \\
    POGARS \cite{thilakarathne2021pose} & Keypoint & Hourglass  & 93.9 & -\\
    COMPOSER \cite{zhou2021composer} & Keypoint & HRNet  & \textbf{94.6} & \textbf{96.2} \\ 
    \midrule
    CRM \cite{azar2019convolutional} & Mixed & I3D & 93.0 & 85.8\\
    TCE+STBiP \cite{yuan2021learning} & Mixed & VGG-16/HRNet  & 94.7 & -\\
    SACRF \cite{pramono2020empowering} & Mixed & I3D/AlphaPose  & 95.0 & 95.2 \\
    GroupFormer \cite{li2021groupformer} & Mixed & I3D/AlphaPose & \textbf{95.7} & \textbf{96.3} \\ \midrule
    \textbf{DECOMPL} & RGB & VGG-16 & \textbf{93.8} & \textbf{95.5} \\ 
    \bottomrule
\end{tabular}
\end{adjustbox}
\vspace{1.5mm}
\caption{Comparisons with SOTAs on the Volleyball dataset with original annotations.}
\label{tables:2}
\end{table}

\subsection{Comparison with the State-of-the-Art (SOTA)}

\noindent \textbf{The Volleyball Dataset}. In order to fairly compare with the previous results reported by the prior work (cf. the SOTA methods in Table \ref{tables:2}), we first evaluated DECOMPL on the original VD (i.e., original annotations). Although we do not exploit temporality, our method is the $2^{\text{nd}}$ best performing (Table \ref{tables:2}) among the RGB-only methods with a $93.8\%$ GAR accuracy. This accuracy is comparable with that of the best performing one, where the difference is in a margin of only $0.3\%$. Moreover, our methods performs also comparably with most of the keypoint only methods and even with certain other mixed methods. Considering the mistakes in the annotations, it would not be reliable to compare the performances based on the original VD. Therefore, we reproduced the results of the compared methods (Table \ref{tables:3}), with publicly available codes \footnote{\href{https://github.com/JacobYuan7/DIN-Group-Activity-Recognition-Benchmark}{https://github.com/JacobYuan7/DIN-Group-Activity-Recognition-Benchmark}}
\footnote{\href{https://github.com/hongluzhou/composer}{https://github.com/hongluzhou/composer}}, with our corrected annotations.

Table \ref{tables:3} shows that DECOMPL achieves an impressive performance of $95.2\%$ GAR accuracy. It is the $2^{\text{nd}}$ highest; yet if we drop the ball tracklets from COMPOSER, our accuracy is the highest among the $4$ compared methods. In particular, our method surpasses DIN by $0.9\%$, which is a prominent RGB-only method, without using any temporal information. Since we could not reproduce the reported results from GroupFormer for mixed inputs, it is excluded from the analysis. Table \ref{tables:2} and Table \ref{tables:3} demonstrate that while the earlier state of the arts DIN and GroupFormer with RGB-only input struggle to benefit from the corrected annotations with only $0.7\%$ and $0.35\%$ respectively, DECOMPL achieves to gain more with a jump of $1.4\%$ in accuracy. Remarkably, all of the methods, in general, benefit from the annotation correction. Overall, we emphasize that only the RGB only methods are comparable to ours, in this respect, our method achieves the second highest performance in Table \ref{tables:2} (original erroneous annotations) and the highest in Table \ref{tables:3} (corrected annotations). 

\begin{table}[t!]
\centering
\begin{tabular}{*3l} \toprule
    Model & Input & Accuracy \\ \midrule
    SACRF \cite{pramono2020empowering} & RGB   & 92.8 \\
    DIN \cite{yuan2021spatio} & RGB   & 94.3 \\
    GroupFormer \cite{li2021groupformer} & RGB   & 94.45 \\
    COMPOSER \cite{zhou2021composer} & Keypoint   & \textbf{96.26} \\ 
    COMPOSER \cite{zhou2021composer} w/o ball & Keypoint   & 94.39 \\ 
    \midrule
    \textbf{DECOMPL} & RGB   & \textbf{95.2} \\ 
\bottomrule

\end{tabular}
\vspace{1.5mm}
\caption{Comparisons with SOTAs on the Volleyball dataset with corrected annotations.}
\label{tables:3}
\end{table}

\begin{table}[t!]
\centering
\begin{tabular}{*3l} \toprule
    Model & $\#$Params & FLOPs \\ \midrule
    ARG \cite{wu2019learning} & 25.182M   &  5.436G \\
    AT \cite{gavrilyuk2020actor} & 5.245M   & 1.260G \\
    HiGCIN \cite{yan2020higcin} & 1.051M  & 184.992G\\
    SACRF \cite{pramono2020empowering} & 29.422M  & 76.757G\\
    DIN \cite{yuan2021spatio} & 1.305M  & 0.311G\\
    COMPOSER \cite{zhou2021composer} & 11.102M  & 0.777G\\
    GroupFormer \cite{li2021groupformer} & 81.52M  & 10.99G\\
    \midrule
    \textbf{DECOMPL} & \textbf{0.65M} & \textbf{0.031G} \\    
    \bottomrule

\end{tabular}
\vspace{1.5mm}
\caption{Computational complexity analysis performed without the backbone and embedding layer.}
\label{tables:6}
\end{table}

\noindent \textbf{The Collective Activity Dataset}. While our primary focus is on the Volleyball dataset, we also conducted experiments on the Collective Activity dataset to demonstrate the effectiveness of DECOMPL. Unlike the Volleyball dataset, CAD does not contain sub-task labels nor a team structure that allows us to split actors in the scene. Therefore, we used a single multi-headed attention block as opposed to two -one for each team- to extract frame features from the dataset. Despite these challenges, DECOMPL achieved a high level of performance with an accuracy of $95.5\%$, which represents a $0.9\%$ improvement in the RGB category and the third-best overall result when keypoint and mixed methods are included. The CAD dataset presents a particular challenge due to the potential ambiguity between the "waiting" and "moving" categories when processing individual frames, as it is not ideal to capture the motion from a single frame. This places a disadvantage for our model in comparison to others exploiting temporality. Nevertheless, our results highlight the power of the attention pooling mechanism for group activity recognition.


\subsection{Computational Complexity Analysis}

In addition to the FLOPS analysis provided by \cite{yuan2021spatio}, for both mixed and keypoint-only categories, we further provide the FLOPS and number of parameters for GroupFormer and COMPOSER, two of the most competitive state-of-the-art methods in these categories in Table \ref{tables:6}. The reported numbers exclude the parameters from the backbone and embedding layer to ensure comparability with prior work. DECOMPL has by far the lowest computational cost by requiring only $0.031$ GFLOPs for a forward pass. Without sacrificing accuracy, it is a remarkable achievement to reduce the number of floating point operations to $10\%$ of the second lowest method. It can be seen that modeling temporality has its costs on the both computational complexity and the number of parameters. DECOMPL is the lightest model in terms of the number of parameters by having only $0.65$ million parameters.



\subsection{Ablation Study}

The results of our ablation study (Table \ref{tables:4} and Table \ref{tables:5}) are obtained by averaging $5$ runs on the validation set of the reannotated VD.

\begin{table}[ht]
\centering
\begin{tabular}{*2l} \toprule
    Ablation & Accuracy \\ \midrule
    only coordinate module & 73.5 \\ 
    w/o coordinate module  & 94.8 \\
    w/o multiple loss signals & 94.7 \\
    max pooling & 94.6 \\
    mean pooling & 94.7 \\
    \textbf{DECOMPL} & \textbf{95.2} \\ \bottomrule
\end{tabular}
\vspace{1.5mm}
\caption{Ablation study on the coordinate module and multiple loss signals.}
\label{tables:4}
\end{table}

Regarding the coordinate module, Table \ref{tables:4} reports the GAR accuracies for the two cases of (i) only the coordinate module and (ii) our method without the coordinate module. As demonstrated, the coordinate module single-handedly achieves $73.5\%$ which is remarkable considering it does not use any visual information. The configuration that players are in contains significant information that should not be overlooked. Moreover, when the coordinate module is not used, the overall performance of our method drops (by $0.4\%$) to $94.8\%$ which is significant as further improvements are more challenging to attain at higher levels. Regarding the use of multiple loss signals (Table \ref{tables:4}), we find that exploiting the decomposable structure of the problem reinforces the representation capacity. The $2$ additional loss signals on top of the group activity and individual activity losses help to increase the accuracy by $0.5\%$. As for the number of heads of the attention pooling, stacking up multiple attention pooling blocks up to $2$ heads is observed in Table \ref{tables:5} to give the best performance. A slight degrade is observed for stacking further up to $4$ and $8$. Finally, two popular permutation invariant pooling techniques are explored. The max pooling is outperformed slightly by the mean pooling, cf. Table \ref{tables:4}. Our results demonstrate the effectiveness of assigning weights to the players in a learnable manner. The attention pooling allows our model to represent the scene better and therefore, an increase of $0.5\%$ in accuracy is observed.

\begin{table}[ht]
\centering
\begin{tabular}{*2l} \toprule
    Heads & Accuracy \\ \midrule
    1  & 94.8 \\
    2 & \textbf{95.2} \\ 
    4 & 94.7 \\ 
    8 & 94.8 \\ \bottomrule
\end{tabular}
\vspace{1.5mm}
\caption{Comparisons of the number of heads in the attention layer.}
\label{tables:5}
\end{table}

\section{Conclusion}
In this paper, we proposed a novel group activity recognition (GAR) technique, DECOMPL, for volleyball videos. DECOMPL effectively complements the visual information with the spatial configuration of the players. Our experiments show that exploiting the problem structure by using multiple auxiliary losses improves the model's representation capacity significantly. We also presented the erroneous annotations on the Volleyball dataset (which is widely used in the literature) and provided the corrected reannoations in a systematic way. Among the state-of-the-art RGB only methods, DECOMPL achieves the best GAR performance with the corrected reannoations and the second best GAR performance with the original annotations.

{\small
\bibliographystyle{ieee_fullname}
\bibliography{egbib}
}

\end{document}